# Adaptive Inventory Strategies using Deep Reinforcement Learning for Dynamic Agri-Food Supply Chains

Amandeep Kaur* and Gyan Prakash

ABV-Indian Institute of Information Technology and Management Gwalior, Madhya Pradesh, India.

*Email: amandeepkaur@iiitm.ac.in; gyan@iiitm.ac.in*

*Abstract:* Agricultural products are often subject to seasonal fluctuations in production and demand. Predicting and managing inventory levels in response to these variations can be challenging, leading to either excess inventory or stockouts. Additionally, the coordination among stakeholders at various level of food supply chain is not considered in the existing body of literature. To bridge these research gaps, this study focuses on inventory management of agri-food products under demand and lead time uncertainties. By implementing effective inventory replenishment policy results in maximize the overall profit throughout the supply chain. However, the complexity of the problem increases due to these uncertainties and shelf-life of the product, that makes challenging to implement traditional approaches to generate optimal set of solutions. Thus, the current study propose a novel Deep Reinforcement Learning (DRL) algorithm that combines the benefits of both value- and policy-based DRL approaches for inventory optimization under uncertainties. The proposed algorithm can incentivize collaboration among stakeholders by aligning their interests and objectives through shared optimization goal of maximizing profitability along the agri-food supply chain while considering perishability, and uncertainty simultaneously. By selecting optimal order quantities with continuous action space, the proposed algorithm effectively addresses the inventory optimization challenges. To rigorously evaluate this algorithm, the empirical data from fresh agricultural products supply chain inventory is considered. Experimental results corroborate the improved performance of the proposed inventory replenishment policy under stochastic demand patterns and lead time scenarios. The research findings hold managerial implications for policymakers to manage the inventory of agricultural products more effectively under uncertainty.

## 1. Introduction

The agri-food supply chain forms complex networks that involve various actors working collaboratively to deliver products and services, from farm to customers [1]. The primary stakeholders who are directly involved in the underlying supply chain are farmers, distributors, retailers, and consumers, and product, information and financial flows among these stakeholders, constitute a complex multi-echelon structure. In 2022, the global market for the agricultural supply chain was US $0.86 billion and is expected to reach US $1.822 billion by 2030, growing at a compound annual growth rate (CAGR) of 10.02% [2]. The agri-food supply chain is crucial for achieving the Sustainable Development Goals (SDGs), particularly SDG 2, which aims to ensure that everyone has access to adequate, safe, and nutritious food. The 2030 Agenda for Sustainable Development, particularly SDG 12, sets a target to halve per-capita global food waste at the retail and consumer levels and to decrease food losses throughout production and supply chains [3].

Although the agri-food supply chain has similarities to other general supply chains, they have specific characteristics such as seasonality, strong perishability, high timeliness and temperature-controlled storage that make their procurement, processing and distribution management quite challenging. Once food products are produced, they need to be processed, stored, delivered, and retailed to meet the diverse demands of customers [4]. Recent reports indicate that about 13% of food is lost between harvest and retail. Furthermore, 17% of the world's food production is wasted at the household, food service, and retail levels [5]. Fruits and vegetables account for approximately 32% of food losses, followed by non-vegetarian products that account for 12.4% of food losses [6]. Hence, implementing effective inventory management strategies is crucial to sell these products before they lose their market value.

Perishability of agri-food products is considered as another important concern that affects overall decisions of agri-food supply chain. Due to perishable nature of many products, seasonality and demand fluctuations poses unique challenges in inventory management of these products. Holding too high inventory will increase procurement costs, and holding costs and wastage costs whereas holding too low inventory results to loss the customers. The products within these chains deteriorate in value as well as in quality once they are produced [7] and also faces increased regulations and environmental pressures, such as packaging, sustainability issues, waste reduction and recycling. Therefore, it is essential to consider the perishability of agri-food products when planning inventory management strategies. This approach helps to reduce inventory costs and enhance the overall efficiency and benefits of the multi-echelon supply chain network.

In a multi-echelon agri-food supply chain, both stochastic replenishment lead times and demand variability significantly impact inventory management, service levels, costs, and coordination efforts [8]. The combined effect of these uncertainties creates a more complex and challenging environment for supply chain managers. The unpredictability of lead times necessitates maintaining higher safety stock to avoid stockouts, which in turn affects reorder points and escalates holding costs [9]. Similarly, fluctuating demand patterns require additional buffer inventory to meet unexpected spikes in customer orders. This dual uncertainty often results in even higher inventory levels than would be necessary if only one factor variable. The interplay between stochastic lead times and demand can lead to more frequent stockouts and backorders, potentially undermining customer satisfaction and resulting in lost sales [10].

Furthermore, this combined uncertainty can increase ordering and transportation costs due to the need for more frequent orders, expedited shipping, or last-minute sourcing from alternative suppliers. The variability in both supply and demand may also necessitate more sophisticated forecasting and inventory optimization techniques, potentially increasing operational complexity and associated costs [11]. The ripple effects of these uncertainties can propagate through the entire multi-echelon structure, amplifying as they move upstream – manifesting the so-called bullwhip effect [12]. This can lead to increased volatility in order quantities at higher echelons, further complicating inventory management and coordination efforts across the supply chain.

In the complex and dynamic environment of agri-food supply chains, a multitude of factors contribute to the complexity of decision-making and profit optimization. These include uncertain demand and supply, fluctuating customer preferences, the perishable nature of food products, seasonal variations, and stringent quality and safety standards. Given this dynamic environment, continuously monitoring and optimizing the entire supply chain for profitability presents a formidable challenge [13]. Effective inventory optimization plays a crucial role in ensuring timely product availability by automatically determining appropriate order quantities. Historically, researchers and practitioners have employed various classical approaches to address these challenges in agri-food supply chain management. Many classical approaches in the agri-food supply chain management are based on game theory [14-17], heuristics [18], control theory [19], Mixed Integer Non-Linear Programming (MINLP) [20], Analytical Hierarchy Process (AHP) and Genetic algorithm [21]. While these traditional solutions have been effective in static environments with specific constraints, they frequently fail to keep pace with the complexities and ever-changing conditions of agri-food supply chains. The limitations of these approaches become apparent when faced with the multifaceted challenges inherent in this sector. The rapidly changing landscape of agri-food supply chains demands more flexible, adaptive, and robust methodologies. These should be capable of handling the inherent uncertainties, incorporating real-time data, and adjusting to shifting market conditions. As such, there is a growing need for innovative approaches that can effectively navigate the complexities of modern agri-food supply chains while maintaining operational efficiency and profitability.

In recent years, the application of Machine Learning (ML) approaches has gained significant traction in addressing the complexities of agri-food supply chain management. These advanced techniques offer promising solutions to improve efficiency, reduce waste, and optimize product procurement processes in highly dynamic environments. Reinforcement Learning (RL) has emerged as a powerful ML method capable of learning optimal strategies without requiring a pre-defined network model. This makes it particularly suitable for the dynamic settings typical of agri-food supply chains. However, the performance of RL algorithms can be limited when dealing with high-dimensional state spaces [22-24]. To overcome these limitations, deep learning is integrated with RL, creating Deep Reinforcement Learning (DRL) in various supply chain optimization scenarios. This hybrid approach provides a robust framework for sequential decision-making in complex stochastic environments [25-26]. Chong et al. [27] proposed a DRL model for apparel supply chain optimization, considering factors such as sell-through rate, service level, and inventory-to-sales ratio. However, their model did not account for uncertain demand scenarios, limiting its real-world applicability. Wu et al. [28] introduced a derivative-free RL approach for multi-echelon supply chain optimization, specifically designed to handle complex stochastic systems. Demizu et al. [29] developed a model-based DRL for inventory management of new smartphones. Their approach incorporated Bayesian neural networks for probabilistic predictions and model-agnostic meta-learning for demand forecasting. Recently, Yavuz and Kaya [30] proposed using DRL algorithms to tackle the dynamic pricing and ordering issues associated with perishable products. They applied Deep Q Learning (DQN) for discrete action spaces and Soft Actor-Critic (SAC) for continuous action spaces in their study. This study offered notable solutions for solving dynamic pricing problems of perishable products based on price and age-dependent stochastic demand. Despite these advancements, challenges persist. The performance of these models can be limited when dealing with large state spaces resulting from stochastic demand, lead times, and product perishability. As the state space grows, the neural networks used in approaches like SAC need to be larger to capture the increasing complexity, leading to higher computational requirements.

Building upon these recent advancements and acknowledging the persistent challenges in considered supply chain management, the work propose a novel approach to address the complexities of inventory optimization under uncertainty. Our research introduces an advanced version of a policy-based Deep DRL algorithm: The Asynchronous Advantage Actor-Critic with Distributed Proximal Policy Optimization (A3C-DPPO). This innovative algorithm is specifically designed to determine optimal inventory policies for agri-food supply chains characterized by stochastic customer demand patterns and variable lead times. The A3C-DPPO approach offers several key advantages. It is well-suited for handling continuous action spaces, allowing for precise determination of optimal ordering quantities for each product in the inventory. This flexibility is crucial in the context of agri-food supply chains, where order quantities often need fine-tuning based on various factors such as perishability and demand fluctuations. Moreover, our proposed technique for replenishment decisions is designed to scale effectively on asymmetric systems, which is particularly valuable in real-world scenarios where retailers within the same supply chain may have different lead times and demand distributions. By incorporating both the A3C and DPPO algorithms, the proposed approach is better equipped to handle the high levels of uncertainty inherent in agri-food supply chains. This includes variability in customer demand, lead times, and other stochastic elements that can impact inventory management. The combination of A3C and DPPO algorithms aims to enhance learning stability and sample efficiency, which are critical factors when dealing with the complex, high-dimensional state spaces typical of agri-food supply chains. Additionally, the distributed nature of the algorithm allows for parallel processing and learning, potentially leading to faster convergence and more robust policies. By addressing the limitations of previous approaches and incorporating advanced DRL techniques, our A3C-DPPO algorithm represents a significant step forward in the application of artificial intelligence to agri-food supply chain management. This approach has the potential to provide more accurate, adaptive, and efficient inventory optimization strategies, ultimately leading to reduced waste, and enhanced overall supply chain performance. In this context, the research contributions of this work are summarized as:

- The study address inventory optimization in a multi-echelon agri-food supply chain network, formulating it as a Markov Decision Process (MDP) to minimize costs and maximize profitability while accounting for stochastic scenarios. To address this complex optimization problem, a novel policy-based Deep Reinforcement Learning algorithm A3C-DPPO is proposed. This advanced algorithm effectively handles continuous action spaces, learning from actions taken within the current policy to determine optimal ordering quantities for each product. Our approach represents a significant advancement in addressing the nuanced decision-making required in inventory management under uncertainty.
- A robust inventory optimization policy capable of adapting to a wide range of demand patterns and replenishment lead time variations. This policy is designed to perform well across various scenarios, enhancing the resilience of the supply chain. It incorporates advanced forecasting techniques and dynamic reorder point calculations to maintain optimal inventory levels despite fluctuations in demand and supply.
- To address potential inventory scarcity during periods of high demand variability, a cooperative framework that operates across multiple stakeholders - including retailers, distribution centers, and agricultural farms is introduced. This distributed approach aims to minimize the impact of uncertainties while simultaneously enhancing the resilience and profitability of the entire agri-food supply chain. The framework facilitates real-time information sharing, collaborative forecasting, and coordinated decision-making among supply chain partners, enabling more efficient resource allocation and risk mitigation strategies.

The article is structured as follows: Section 2 provides a literature review to offer a comprehensive overview of existing research. The subsequent section covers the problem formulation. In Section 4, novel A3C-DPPO-based intelligent inventory management framework is presented. Section 5 presents numerical analyses of the formulated problem. Section 6 discusses theoretical contributions and managerial implications arising from the research. Finally, Section 7 offers concluding remarks and suggestions for future research directions.

2. **Related Work**

To comprehensively investigate the existing research domain, a thorough review of the literature is conducted. This section explores both classical approaches applied in the food supply chain context to capture broader operational aspects, and reinforcement learning (RL)-based methods used in supply chain optimization to provide valuable insights into the current state of knowledge.

*2.1 Classical Approaches*

Inventory management is crucial for optimizing operations, meeting customer demands, reducing costs, and maintaining the quality of agri-food products throughout the supply chain. Effective inventory management results in efficient resource utilization by minimizing overstocking and reducing waste, leading to cost savings and improved profitability. In recent years, comprehensive studies on the agri-food supply chain with classical approaches have been discussed in the literature. For instance, the authors in [31] proposed ordering and pricing policy for deteriorating items having Weibull survival and death characteristics. The authors considered continuous stochastic demands with an objective to maximize the system's total average expected profit based on a continuous review *(r, Q)* policy. To solve the complex stochastic differential equations, direct approach-Taylor series expansion was adopted to obtain an optimal strategy. In [32], the authors formulated a single-stage and two-stage pricing and inventory decision model for retailers dealing with fresh agricultural products. Liu *et al.* [33] addressed optimal purchase and inventory replenishment decisions for perishable agricultural products, considering storage costs, inventory shortages and surpluses, price and demand fluctuations, and product deterioration. They focused on a finite-period inventory model for wholesalers where demand is influenced by current and past prices. Sindhuja *et al.* [34] developed an inventory model for deteriorating dairy products, incorporating quality-based demand to mitigate deterioration rates and minimize total incurred expenditures and costs. Additionally, Banerjee and Agrawal [35] devised an inventory model for deteriorating items, considering demand dynamics influenced initially by selling price and subsequently by freshness condition.

Several researchers have proposed various solution approaches for optimizing agri-food supply chains, including game theory, fuzzy Analytic Hierarchy Process (AHP), Linear Programming (LP), Mixed Integer Linear Programming (MILP), Mixed Integer Programming (MIP), MILNP, Genetic Algorithm (GA), Multi-objective Genetic Algorithm-II (MOGA-II), Non-Sorted Genetic Algorithm (NSGA), and Particle Swarm Optimization (PSO). Validi *et al.* [36] introduced a robust solution using Technique for Order of Preference by Similarity to Ideal Solution (TOPSIS) for a two-layer dairy supply chain involved in milk distribution in Ireland. They applied TOPSIS to rank transportation routes based on trade-offs between total costs and $CO_2$ emissions. In addition, Mirakhorli *et al.* [37] utilized the Fuzzy Multi-Objective Linear Programming (FMOLP) method to optimize the supply chain of a bread-producing factory, aiming to achieve Pareto-optimal solutions that minimize total costs and delivery time simultaneously. Galal *et al.* [38] studied a two-echelon agri-food supply chain, exploring the impact of varying order quantities on costs, emissions, and service levels. They developed a discrete-event simulation model to account for stochastic demand and lead-time variability, as well as effects on service levels and product lifetimes across the supply chain. Miranda *et al.* [39] modelled and optimized green supply chain with multi-objective optimization via GA and TOPSIS and then validated through the development and analysis of multi-echelon supply chain for orange juice. However, these models did not consider the uncertainty in supply chain as a primary issue. To address this issue, Zhao and Wang [40] proposed an optimization model with GA for inventory control under uncertainty. Further, Mousavi *et al.* [41] proposed modified PSO to solve inventory control problems in two-echelon supply chain as MINLP optimization problems. While both these meta-heuristic optimization techniques can be effective in high-dimensional and complex search spaces, they might less suitable for certain scenarios, such as inventory optimization under uncertain demand. Both these algorithms may struggle to find robust solutions that perform well under various demand scenarios.

The current literature on classical inventory management approaches often lacks comprehensive integration of innovative technologies into supply chain practices. In the agri-food supply chain specifically, there is a noticeable gap in studies that explore inventory management under uncertainties and their broader impacts on supply chain dynamics.

*2.2 RL based Approaches*

Inventory demand patterns can exhibit non-stationarity over time due to seasonal fluctuations, market trends, or unforeseen events. RL based approaches can adapt to non-stationary environments by continuously updating their policies based on recent experiences, allowing them to maintain performance in dynamically changing demand scenarios. Basically, RL and its variants formulates the problem with Markov Decision Process (MDP) which does not requires explicit mathematical model to develop optimal policy even under highly complex scenarios. In supply chain scenarios, RL techniques have been explored to optimize various aspects such as ordering decisions [25], inventory management of perishable products [24], and procurement and distribution functions. Kara *et al.* focused on ordering policies for perishable inventory items within a multi-retailer framework, formulating the problem using RL temporal-difference algorithms, specifically Q-Learning and State-Action-Reward-(next)State-(next)Action (SARSA). A similar learning model in multi-retailer competitive environment was proposed by Dogan and Güner [22] based on stochastic dynamic programming and agent-based simulations. Regrettably, such algorithms can only handle problem with limited state-action space, has initiated many studies toward DRL based solutions for inventory management in highly complex supply chain network.

Recently, Boute *et al.* [26] described the key design choices of DRL algorithms in inventory control. Another study [42] proposed DRL solution for lot sizing problem of perishable material under uncertain environment but limits its performance to single supplier setting. Being DRL architecture was introduced to address the problem of multi-dimensional action spaces, but it still requires readjustment of network architecture as well as training algorithm for continuous action spaces. Keeping similar flavour of DRL but with different classes of DRL known as Policy Gradient are adopted to handle multi-echelon divergent system under stochastic supply chain environment. In particular, Proximal Policy Optimization (PPO) has proven to be a robust algorithm for large and continuous action space under stochastic multi-echelon supply chain optimization [43]. Research by Geevers *et al.* [44] investigated the performance of PPO algorithm linear, divergent and general structure of multi-echelon supply chain network but limit their performance to specific scenarios only. Hubbs *et al.* [45] developed PPO based dynamic scheduling for chemical production but limits its performance to single stage supply chain networks. Additionally, Tian *et al.* [46] proposed A2C with PPO and GRU attention mechanism for inventory replenishment of warehouse.

Despite the growing adoption of DRL in addressing various supply chain challenges, including inventory management, production planning, quality control, and sustainability improvement, there remains a significant gap in the literature regarding its application to the complex agri-food supply chain. Our comprehensive review of the existing literature reveals a notable scarcity of research that thoroughly explores advanced RL models in the context of agri-food supply chains. While DRL has shown promise in other supply chain domains, its potential in addressing the specific challenges of agri-food systems remains largely untapped.

## 3. Problem Description

### 3.1 Problem Formulation

In the multi-echelon agri-food supply chain network, three stakeholders, which are agricultural farms, distribution centres, and retailers that forms a complex network are considered as illustrated in Fig.1. Farmers produce fresh agricultural products in farms and sell them to distribution centers. The distribution centers sell them further to retailers. As the products considered are perishable having short shelf life, it is assumed that all the products with different remaining shelf-lives are merged and shipped together from the distribution centres to the retailers at the beginning of each period. The distribution centers receive orders from retailers and are mainly responsible for order processing, replenishing inventory of fresh agricultural products procured from farmers, and transporting and distributing them along upstream and downstream supply chains. Under high demand variance and unpredictable lead time, it is necessary to make inventory decisions to optimize the associated inventory costs, such as holding cost, shortage cost, and transportation cost in order to maximise the revenue of the overall supply chain network. Considering this scenario, the inventory optimization model has been developed. The main notations- parameters and variables used in the model formulation are summarized in Table 1.

In order to build the model of a three-echelon supply chain network for fresh agricultural products, few assumptions are made as follows:

- There is finite number of stakeholders in each echelon
- A limited number of transportation vehicles are available
- The flow of products between the entities of same echelon is not allowed
- Each retailer can place order from single distribution centre at particular replenishment interval

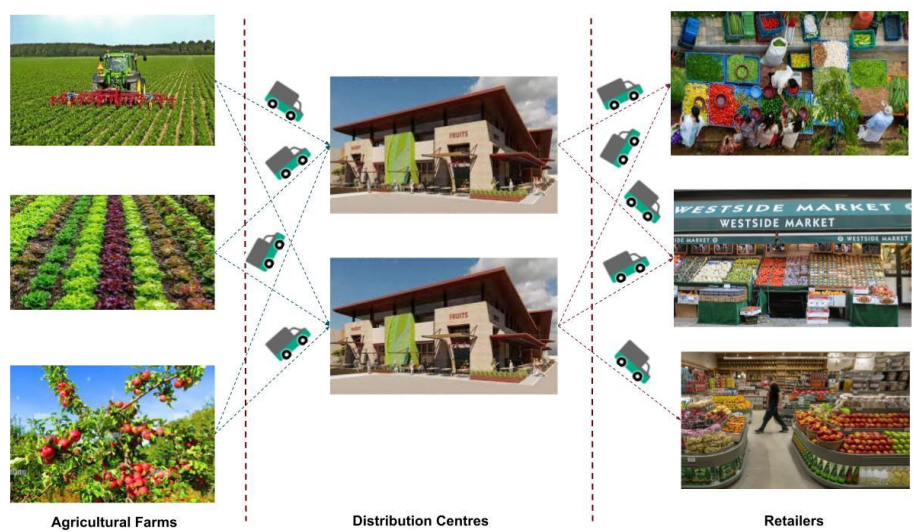

**Fig. 1** Multi-echelon agri-food supply chain network

Table 1. Main Notations

| Notation | Meaning |
|---|---|
| **Sets** | |
| $p \in P$ | Set of crops harvested in agricultural farms |
| $j \in F$ | Set of location of agricultural farms |
| $k \in K$ | Set of distribution centers |
| $m \in M$ | Set of vehicles |
| $t \in T$ | Set of time periods |
| $c \in C$ | Set of retailers |
| **Parameters** | |
| $c_{p,j,t}$ | Purchasing cost associated with product $p$ from agricultural farm $j$ in time period $t$ |
| $c'_{p,c,k}$ | Purchasing cost associated with product $p$ from distribution center $k$ in time period $t$ |
| $c_f$ | Fixed purchasing cost associated with product $p$ |
| $c_{h,p,k}$ | Holding cost associated with product $p$ by distribution center $k$ in time period $t$ |
| $c_{h,p,c}$ | Holding cost associated with product $p$ by retailer $c$ in time period $t$ |
| $v_{t,p,k}$ | On hand inventory of product $p$ available at distribution centers $k$ at time period $t$ |
| $v_{t,p,c}$ | On hand inventory of product $p$ available at retailer $c$ at time period $t$ |
| $a_{t,p,k}$ | Open Order of product $p$ reach at distribution centers $k$ at the beginning of time period $t$ |
| $a_{t,p,c}$ | Open Order of product $p$ reach at retailer $c$ at the beginning of time period $t$ |
| $d_{t,p,c}$ | Demand of product $p$ by the retailer $c$ during time period $t$ |
| $d_{t,p,i}$ | Demand of product $p$ at the retailer $c$ by user $i$ during time period $t$ |
| $c_{w,p,k}$ | Wastage cost associated with product $p$ at distribution centers $k$ |
| $c_{w,p,c}$ | Wastage cost associated with product $p$ at retailer $c$ |
| $c_{sh,p,k}$ | Shortage cost associated with product $p$ at distribution centers $k$ |
| $c_{sh,p,c}$ | Shortage cost associated with product $p$ at retailer $c$ |
| $c^m_{p,l,j}$ | Cost of loading product $p$ in vehicle $m$ to deliver at farm $j$ |
| $c^m_{p,l,k}$ | Cost of loading product $p$ in vehicle $m$ to deliver at distribution centers $k$ |
| $c^m_{p,un,k}$ | Cost of unloading product $p$ from vehicle $m$ to deliver at distribution centers $k$ |
| $c^m_{p,un,c}$ | Cost of unloading product $p$ from vehicle $m$ to deliver at retailer $c$ |
| $dist_{j,k}$ | Distance between farm $j$ and distribution centers $k$ |
| $dist_{k,c}$ | Distance between distribution centers $k$ and retailer $c$ |
| $c_{fuel}$ | Cost (per liter) of fuel usage for transportation from farm location to distribution center |
| $c_{f_k,trans}$ | Fixed cost associated during transportation for distribution center |
| $c_{f_c,trans}$ | Fixed cost associated during transportation for retailer |
| $V_{k,\max}$ | Maximum inventory capacity at distribution center |
| $V_{c,\max}$ | Maximum inventory capacity at retailer end |
| $Q_m$ | Maximum capacity of vehicle $m$ |
| **Variables** | |
| $q_{j,p,k,t}$ | Quantity of product $p$ purchased from farm $j$ by distribution center $k$ in time period $t$ |
| $q_{t,c,p,k}$ | Quantity of product $p$ purchased from distribution center $k$ by retailer $c$ in time period $t$ |
| $I(q_{j,p,k,t})$ | Binary indicator with 1 if order of product $p$ placed from farm $j$ by distribution center $k$ in time period $t$ |
| $q_{h,p,k,t}$ | Quantity of product $p$ to hold at distribution center $k$ in time period $t$ |
| $q_{h,p,c,t}$ | Quantity of product $p$ to hold at retailer $c$ in time period $t$ |
| $q_{w,j,k,t,p}$ | Quantity of product $p$ results into wastage during transportation and handling from farm $j$ to distribution center $k$ in time period $t$ |
| $q_{w,j,c,t,p}$ | Quantity of product $p$ results into wastage during transportation and handling from distribution center $k$ to retailer $c$ in time period $t$ |
| $q_{sh,k,t,p}$ | Shortage quantity of product $p$ at distribution center $k$ in time period $t$ |
| $q_{sh,c,t,p}$ | Shortage quantity of product $p$ at retailer $c$ in time period $t$ |
| $q^m_{p,j,t}$ | Quantity of product $p$ loaded into transportation vehicle $m$ at farm in time period $t$ |
| $q^m_{p,k,t}$ | Quantity of product $p$ loaded into transportation vehicle $m$ at distribution center in time period $t$ |
| $N_{m,j,k,t}$ | Number of vehicles needed to deliver the products from farm location and distribution centers $k$ |
| $N_{m,k,c,t}$ | Number of vehicles needed to deliver the products from distribution centers $k$ to retailer $c$ |
| $l$ | Replenishment lead time (in days) at distribution center |
| $l'$ | Replenishment lead time (in days) at retailer |

The inventory flow starts with distribution centre raising its inventory level by procuring from agricultural farmers at the beginning of each day. Let $k = \{1, 2, \ldots, K\}$ be the distribution centers which procure orders of products from the farmers. During the inventory procurement, purchasing cost, holding cost, delivery cost (or transportation cost), wastage cost, shortage and stockout cost are considered and the objective function is to determine the optimal order quantities to maximize the total revenues and inventory turnover while minimizing inventory related cost and stockouts,

$$\min \left( C_p^t + C_h^t + C_{Trans}^t + C_W^t + C_s^t \right) \tag{1}$$

$$C_p^t = \sum_k \sum_t \sum_j \sum_p c_{p,j,t} \cdot q_{t,j,p,k} + c_f \cdot I(q_{t,j,p,k}) + \sum_k \sum_t \sum_c \sum_p c'_{p,c,k} \cdot q_{t,c,p,k} \tag{2}$$

$$C_h^t = \sum_t \sum_k \sum_j \sum_p c_{h,p,k} \cdot q_{h,k,t,p} + \sum_t \sum_k \sum_c \sum_p c_{h,p,c} \cdot q_{h,c,t,p} \tag{3}$$

where $q_{h,k,t,p}$ and $q_{h,c,t,p}$ represents the quantity of each product category to be hold in the inventory at the distribution center $k$ and retailer end $c$, calculated as

$$q_{h,k,t,p} = \{v_{t,p,k} + a_{t,p,k} - d_{t,p,c}\}^+ \tag{4a}$$

$$q_{h,c,t,p} = \{v_{t,p,c} + a_{t,p,c} - d_{t,p,i}\}^+ \tag{4b}$$

where $a_{t,p,k}$ and $a_{t,p,c}$ represents the open orders received at the beginning of time period $t$ at distribution center $k$ and retailer $c$.

Further, the wastage cost of agricultural products includes economic losses incurred due to the inefficient use or disposal of agricultural products throughout the supply chain. It includes physical loss of agricultural products due to spoilage, damage during handling and transportation. The degradation of these fresh agricultural products also depends on the freshness, can be calculated as $q_{\delta_t} = q_{\delta_p} \cdot e^{-\mu_p \cdot F_{p,t}}$. Here $\delta_p$ indicates the wastage penalty coefficient specific to agricultural product $p$, $\mu_p$ is sensitive factor related to deterioration of fresh agricultural product $p$ with respect to time, $\mu_p > 0$, and $F_{p,t} \in [0,1]$ denotes the freshness level of product $p$ at time $t$, where $F_{p,t} = 1$ indicates a freshly procured item and $F_{p,t} = 0$ represents complete spoilage. The wastage cost increases as the freshness decays and can be calculated as

$$C_w^t = \sum_t \sum_k \sum_j \sum_p c_{w,p,k} \cdot (q_{w,j,k,t,p} + q_{\lambda_t}) + \sum_t \sum_k \sum_c \sum_p c_{w,p,c} \cdot (q_{w,k,c,t,p} + q_{\lambda_t}) \tag{5}$$

The values of $\delta_p$ and $\mu_p$ are product specific represent the perishability profile of each product. For highly perishable products such as leafy vegetables, $\delta_p = 0.18$ and $\mu_p = 6.0$ are considered whereas for low-perishability items, $\delta_p = 0.05$ and $\mu_p = 2.0$ are considered

Additionally, shortage cost refers to the economic losses incurred when there is an insufficient supply of agricultural products to meet the demand due to various factors such as fluctuations in market demand, adverse weather conditions or disruptions in transportation. In monetary terms, it is calculated as

$$C_{sh}^t = \sum_t \sum_k \sum_p c_{sh,p,k} \cdot q_{sh,t,p,k} + \sum_t \sum_c \sum_p c_{sh,p,c} \cdot q_{sh,t,p,c} \tag{6}$$

where $q_{sh,t,p,k}$ and $q_{sh,t,p,c}$ be the shortage in product quantity at distribution center and retailer end respectively, which is calculated as $q_{sh,t,k,p} = \{d_{t,k,p} - v_{t,k,p} - a_{t,k,p}\}^+$ and $q_{sh,t,c,p} = \{d_{t,c,p} - v_{t,c,p} - a_{t,c,p}\}^+$

Finally, the transportation of fresh products from farms to the distribution centres cum warehouses and retailers involves certain expenses, which is calculated as

$$C_{Trans}^t = \left\{ \sum_t \sum_p \sum_{j \in F} \sum_{k \in K} \sum_{m \in M} \left[ \left( \frac{q_{p,j,t}^m}{Q_m} \right) \cdot (c_{p,l,j}^m + c_{p,un,k}^m) \cdot dist_{j,k} \cdot c_{fuel} \right] N_{m,j,k,t} + c_{f_k,trans} \right\} \\ + \left\{ \sum_t \sum_p \sum_{k \in K} \sum_{c \in C} \sum_{m \in M} \left[ \left( \frac{q_{p,k,t}^m}{Q_m} \right) \cdot (c_{p,l,k}^m + c_{p,un,c}^m) \cdot dist_{k,c} \cdot c_{fuel} \right] N_{m,k,c,t} + c_{f_c,trans} \right\} \tag{7}$$

where $c_{f_k,trans}$ and $c_{f_c,trans}$ represents fixed costs as additional expenses associated with transportation from farm location to distribution centers and retailers' end such as toll taxes, insurance and permits.

Let $b(\tilde{t})$ be the binary value, where $b(\tilde{t}) = 1$ indicates the order placed at time period $\tilde{t} < t$ as open order and $b(\tilde{t}) = 0$ otherwise. There is strictly positive replenishment lead time $l$ at distribution centre with associated open orders represented as

$$a_{t,p,j,k} = \left(b(t-l).q_{p,j}(t-l),........,b(t).q_{p,j}(t)\right), \forall p \in P, j \in F \quad (8)$$

Similarly, replenishment lead time $l'$ at retailer end associated with open orders represented as

$$a_{t,p,k,c} = \left(b(t-l').q_{p,k}(t-l'),........,b(t).q_{p,k}(t)\right), \forall p \in P, k \in K \quad (9)$$

The order fulfilment process follows a First-In-First-Out (FIFO)-based order pipeline. Each placed order is associated with a lead-time counter that decrements at each time step. In next time slot $(t+1)$, the inventory levels of distribution centre as well as retailer end are updated as

$$I_k(t+1) = I_k(t) + q_{p,k,t} - \sum_{c=1}^{C} q_{p,c,t}, \forall k \in K \quad (10)$$

$$I_c(t+1) = I_c(t) + q_{c,p,t} - \tilde{d}_c(t+1), \forall c \in C \quad (11)$$

To capture the inventory level, we consider $v(t, \lambda_t)$ for the overview of product available in inventory along with their freshness level corresponding to each inventory product and concatenate them together for distribution center as well as for retailer as $I_k(t) = I_c(t) = \left[v(t, \lambda_o), ....., v(t, \lambda_{t'}), ....., v(t, \lambda_t)\right]$. Specifically, the objective is to minimize the total inventory related cost over all periods subject to the following constraints as

$$\min \sum_t \left[ \underbrace{\sum_k \sum_t \sum_j \sum_p c_{p,j,t}.q_{t,j,p,k} + c_f.I(q_{t,j,p,k}) + \sum_k \sum_t \sum_c \sum_p c'_{p,c,k}.q_{t,c,p,k}}_{\text{Purchasing cost}(C_p)} + \underbrace{\sum_t \sum_k \sum_j \sum_p c_{h,p,k}.q_{h,k,t,p} + \sum_t \sum_k \sum_c \sum_p c_{h,p,c}.q_{h,c,t,p}}_{\text{Holding Cost}(C_h)} \right.$$
$$\left. + \underbrace{\sum_t \sum_k \sum_j \sum_p c_{w,p,k}.(q_{w,j,k,t,p} + q_{\lambda_t}) + \sum_t \sum_k \sum_c \sum_p c_{w,p,c}.(q_{w,k,c,t,p} + q_{\lambda_{t'}})}_{\text{Wastage Cost}(C_w)} + \underbrace{\sum_t \sum_k \sum_p c_{sh,p,k}.q_{sh,t,p,k} + \sum_t \sum_c \sum_p c_{sh,p,c}.q_{sh,t,p,c}}_{\text{Shortage Cost}(C_{sh})} + C_{Trans} \right]$$

(12)

$$C1: v_k(t) \leq V_{k,\max}, \forall k \in K \quad (12a)$$

$$C2: v_c(t) \leq V_{c,\max}, \forall c \in C \quad (12b)$$

$$C3: \sum_{j \in F} I(q(t,j)) \leq 1, \forall t \in T \quad (12c)$$

$$C4: \sum_{k \in K} I(q_c(t,k)) \leq 1, \forall t \in T \quad (12d)$$

$$C5: N_{m,j,t} \leq M, \forall m \in M, j \in F, t \in T \quad (12e)$$

$$C6: \sum_{\substack{j \in F, \\ k \in K, \\ p \in P}} Q_{p,j,k,t}^m \leq Q_m, \forall m \in M, t \in T \quad (12f)$$

Constraint C1 and C2 set limits on inventory items that should be less than or equal to the maximum inventory holding capacity of distribution center and retailer, respectively. For simplicity of the model, constraints C3 and C4 assume that each distribution center and retailer can place orders from a single agricultural farm and a single distribution center, respectively, during each time period $t$. Further, Constraint C5 ensures that the total number of transportation vehicles must not exceed the maximum allowed number. Constraint C6 examines the maximum capacity of each transportation vehicle and prevent its overloading.

### 3.2 Modeling of Demand and Replenishment Lead Time in Stochastic Environment

The agri-food supply chain operates in a highly stochastic environment characterized by numerous sources of uncertainty and variability. These include fluctuating customer demand patterns influenced by factors such as seasonality, weather conditions, and changing consumer preferences; and unpredictable replenishment lead times affected by transportation delays and supply disruptions. The perishable nature of agri-food products introduces additional complexity, as product quality and shelf life can vary significantly. This multifaceted stochastic environment necessitates sophisticated inventory management strategies that can adapt to rapidly changing conditions, balance multiple competing objectives, and make optimal decisions in the face of incomplete information and uncertain outcomes. The stochastic nature of customer demand and replenishment lead times is modeled using a diverse set of probability distributions: specifically, the Normal $(N(\mu,\sigma^2))$, Gamma $(Gamma(\alpha,\beta))$, Weibull $(Weibull(u,\lambda))$, and Exponential $(Exp(\lambda))$ distributions. This approach allows us to capture a wide range of demand and lead time patterns and uncertainties commonly observed in the agri-food supply chains. The Normal distribution is employed to represent symmetric demand fluctuations around a mean, while the Gamma distribution is utilized to model positively skewed demand scenarios often seen in seasonal products. The Weibull distribution, known for its flexibility, is used to capture more complex demand patterns, including those with varying degrees of skewness and kurtosis. For practical lead time scenarios, the Exponential distribution is considered. By incorporating these distinct distributions, our proposed inventory optimization model develops robust and adaptable framework to various real-world scenarios, enhancing its applicability across different product categories and market conditions in the agri-food sector.

## 4. DRL based Intelligent Inventory Management Framework

The cost optimization of the agri-food supply chain network involves complex interactions, multiple decision points, uncertain demands and supply dynamics that can be mapped with MDP. It provides a robust framework for modelling and solving decision-making problems under uncertainty and over time.

### A. MDP Model

In the context of inventory management, the goal is to optimize profits across the agri-food supply chain by minimizing holding costs, ordering costs, and stockouts. The decisions in supply chain often made sequentially over time. MDPs inherently capture the concept of sequential decision-making that is crucial where actions at one echelon impact downstream and upstream operations. This work considers a network consisting of a single agricultural farm, a single distribution center, and multiple retailers. However, the network can be further extended by adding more distribution centers and retailers as sub-network in underlying supply chain. The key components of MDP includes state, action, reward that are defined as follows,

- **State:** The state captures the essential information about the inventory status in time slot $t$. The state includes the information about (1) the inventory level at the distribution center and retailer, (2) replenishment lead time and (3) customer demands

$$x_t = \left\{ I_{c,t}, I_{k,t}, \tilde{l}_k, \tilde{l}_c, \tilde{d}_{c,t}, \varsigma_{t-n} \right\}, \forall k \in K, c \in C \tag{13}$$

where $I_{k,t}$ and $I_{c,t}$ represents the current inventory position of distribution centre and retailers respectively, $\tilde{l}_k$ and $\tilde{l}_c$ indicates the predicted replenishment lead time for distribution center and retailer respectively, $\tilde{d}_{c,t}$ represents the predicted demand by end customers at retailer end and $\varsigma_{t-n}, n \in \{1,2,......,7\}$ represents 7-dimensional vector corresponding to the days of the week. In a stochastic environment, uncertainty in customer demand and lead time is incorporated. To capture demand variability, Normal, Gamma, and Weibull distributions are employed. For practical modeling of lead time uncertainty, Exponential and Gamma distributions are used

- **Action:** The action space considers the order quantity corresponding to distribution centre and retailers as

$$y_t = \{q_{k,t}, q_{c,t}\}, \forall k \in K, c \in C \tag{14}$$

Here, the action space is considered continuous ordering quantities. The continuous action space allows for greater flexibility in optimizing order size to meet specific objectives while minimizing inventory related costs, avoiding stockouts or balancing inventory levels.

- **Reward:** The reward obtained by each agent after executing an action $y_t$ in particular state $x_t$. The net profit is calculated by considering revenue generated at individual entity in underlying supply chain and inventory cost incurred at retailer and distribution center as

$$z_t = \begin{cases} R_k(t) + \sum_{c \in C} R_c(t) - \left( C_k(t) + \sum_{c \in C} C_c(t) \right), & \text{if C1 to C6 are satified} \\ 0, & \text{otherwise} \end{cases} \tag{15}$$

The reward calculates the average revenue generated in entire supply chain that can be calculated as

$$z(t+1) = \sum_{t=1}^{T} z_t \tag{16}$$

The value function $V^\pi$ used to map strategy $\pi$ in particular state $x_t$ as

$$\begin{aligned} V^\pi(x(t), y(t)) &= \mathbb{E}[z(t) | x(t), y(t)] \\ &= \mathbb{E}[z(t) + \gamma z(t+1) + \ldots | x(t), y(t)] \\ &= \mathbb{E}[z(t) + \gamma V^\pi(x(t+1), y(t+1)) | x(t), y(t)] \end{aligned} \tag{17}$$

where $\gamma \in [0,1]$ represents the discount factor, reflecting the significance of future rewards. The total expected discounted reward with optimal strategy under Bellman optimal equation as

$$\pi^* = \arg\max_{y} V^\pi(x, y) \tag{18}$$

*B. DQN Algorithm*

The DQN algorithm obtains an optimal inventory replenishment policy by approximating the value function with a neural network. It stores the information in the form of weight updates corresponding to each state pair in the neural network. In this considered scenario, the environment is continuously observed the state space information to decide on the optimal order quantity for maintaining an efficient supply chain. The DQN algorithm uses main network and target network to approximate the optimal action-value function $Q(x, y)$, which estimates the expected cumulative reward $z_t$ for taking an action $y$ in a given state $x$. The main network is responsible for selecting actions, while the target network provides stable target values for learning. The action-value function is updates using the Bellman equation

$$Q(x, y) = z_t + \gamma \max_{y'} Q(x', y') \tag{19}$$

where $x' = x(t+1)$ represents next state. The learning process involves storing experiences $(x, y, z, x')$ in an experience replay buffer. During training, mini-batches of experiences are sampled from the buffer to update main network. The loss function $L(\theta)$, which measures the difference between predicted and target values, is minimized using gradient descent as

$$L(\theta) = \mathbb{E}_{(x,y,z,x')} \left[ \left( z + \gamma \max_{y'} Q(x', y'; \theta^-) - Q(x, y; \theta) \right)^2 \right] \tag{20}$$

where $\theta$ and $\theta^-$ are the parameters of the main network and target network respectively.

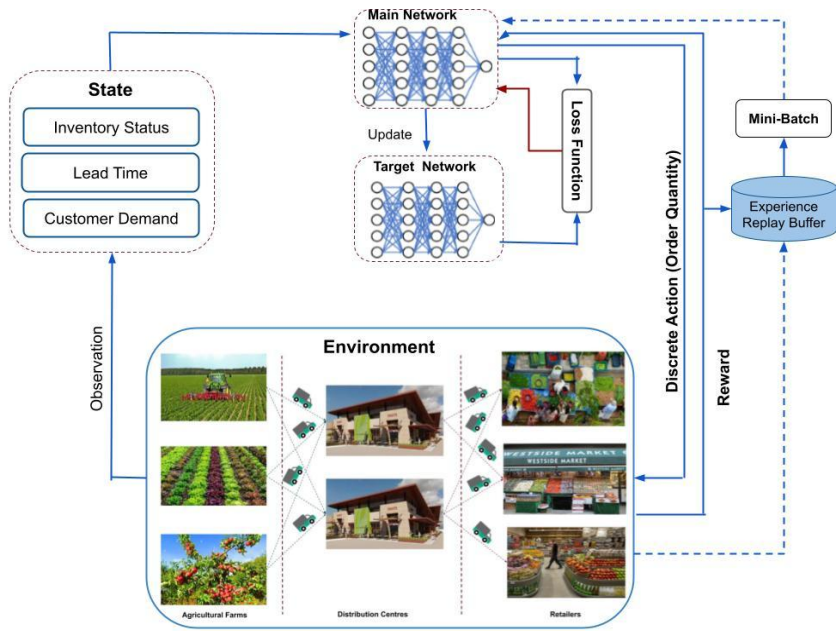

Fig. 2 Inventory optimization with DQN Algorithm

The discrete actions, represented as order quantities are chosen to optimize the overall reward, which is a function of factors such as cost minimization and demand satisfaction. By iteratively updating the networks and refining its predictions, the DRL framework learns to manage the inventory effectively, ensuring that customer demand is met while minimizing waste and holding costs.

*C. Soft Actor-Critic (SAC) Algorithm*

The SAC algorithm is designed to handle continuous action space in stochastic environment. Basically, SAC combines the benefits of both value-based and policy-based RL approaches. The algorithm is designed to maximize a trade-off between the expected return and the entropy of the policy. The entropy term encourages exploration by preventing the policy from becoming too deterministic. The SAC algorithm consists of an actor (policy) network and two critic (value) networks, and a value function, each parameterized by neural networks.

The overall objective of SAC is to maximize the expected return while incorporating an entropy during exploration

$$J(\pi) = \sum_{t=0}^{T} \mathbb{E}_{(x_t, y_t) \sim \rho_\pi} \left[ z(x_t, y_t) + \delta H(\pi(\cdot | x_t)) \right] \tag{21}$$

where $\delta$ is the temperature parameter that balances reward maximization and entropy, and $H(\pi(\cdot | x_t)) = -\log \pi(y_t | x_t)$ indicates the entropy of the policy. The SAC uses two Q-functions $Q_{\theta_1}(x, y)$ and $Q_{\theta_2}(x, y)$ to mitigate positive bias in the value estimates. The value function

$$V_\psi(x) = \mathbb{E}_{y \sim \pi_\phi} \left[ Q_{\theta_1}(x, y) - \delta \log \pi_\phi(y | x) \right] \tag{22}$$

The loss function for updating the value network is

$$J_V(\psi) = \mathbb{E}_{x \sim D} \left[ \left( V_\psi(x) - \mathbb{E}_{y \sim \pi_\phi} \left[ Q_{\theta_1}(x, y) - \delta \log \pi_\phi(y | x) \right] \right)^2 \right] \tag{23}$$

The policy $\pi_\phi(y | x)$ is updated to maximize the expected return and entropy

$$J_\pi(\phi) = \mathbb{E}_{x \sim D, y \sim \pi_\phi} \left[ \delta \log \pi_\phi(y | x) - Q_{\theta_1}(x, y) \right] \tag{24}$$

This encourages the policy to take actions that not only maximize the Q-value but also maintain high entropy. By integrating entropy maximization into the RL objective, SAC ensures a balanced trade-off between exploration and exploitation. The use of dual Q-functions, a value function, and a stochastic policy network, all updated with specific loss functions, provides the algorithm with the robustness needed to handle complex and uncertain environments efficiently.

*D. Proposed Cooperative A3C-DPPO Algorithm*

To handle the continuous action space required for controlling product order quantities and enabling optimal decision-making, the A3C-DPPO algorithm is adopted. The algorithm executes its training process in distributed manner and follows similar training as Asynchronous Advantage Actor-Critic (A3C) multi-agent learning (Zhang *et al.*, 2019). Like A3C, DPPO has multiple learning agents placed in different locations that interacts independently with the environment, conduct model training and periodically update the global PPO network by collecting input from the local network as gradient of each agent. In the considered agri-food supply chain, each retailer (act as local agent) is connected to a single distribution center (act as global model), that is updated by collecting information from the local models and synchronize global parameters to local parameters, follows the training process of A3C-DPPO algorithm. Similarly, each distribution center is connected to single agricultural farm. A similar distributed training framework is adopted for the inventory model of the distribution center, wherein the supplier (agricultural farm) functions as the global agent and the distribution centers serve as local agents to enable collaboration among stakeholders. The global model updates the corresponding parameters by collecting the gradients of each agent and sends the updated model to each local agent for next round of training. In distributed setting, PPO can coordinate with A3C to utilize the information gathered by multiple agents efficiently while maintaining policy stability. The cooperative aspect arises from the synergy between A3Cs exploration and PPOs optimization. Together they work cooperatively to improve the policy and value function toward achieving desired objective. The proposed A3C-DPPO architecture introduces a cooperative learning framework in a hierarchical supply chain network. The retailers act as local agents interacting with a distribution center. The distribution centers act as global agents aggregating feedback from multiple retailers. Each agent maintains a local actor-critic network, which learns policies and value functions independently based on its environment interactions. These local gradients are periodically aggregated at their corresponding global model (distribution center), which then updates shared parameters and redistributes them to local agents.

The actor network learns the policy $\pi_{\theta_k}(u_k(t)|v_k(t))$ which determines the probability distribution over actions given the current state $u_k(t)$. The critic network estimates the value function $\vartheta_{\phi_k}(u_k(t))$, which evaluates the expected return from a given state $u_k(t)$. The policy update for the retailer $k$ aims to maximize the expected reward by adjusting the policy parameters $\theta_k$ as

$$\theta_k(t+1) = \theta_k(t) + \alpha_a \nabla_{\theta_k} \log \pi_{\theta_k}(v_k(t)|u_k(t)) A_k^\pi \tag{25}$$

where $\alpha_a$ is the learning rate for the actor network and $\nabla_{\theta_k} \log \pi_{\theta_k}(v_k(t)|u_k(t))$ is the gradient of the log-probability of the action taken. To update the weights of the actor network, the advantage of taking a certain action over other actions need to be calculated such that the policy is updated in the direction that increases the chances of taking better actions. The advantage function $A_k^\pi$ corresponding to each retailer is defined as

$$A_k^\pi = \left(r_k(t) + \gamma \vartheta_{\phi_k}(u_k(t+1)) - \vartheta_{\phi_k}(u_k(t))\right) \tag{26}$$

The critic network update adjusts the value function parameters $\phi_k$ to minimize the difference between the predicted and actual returns as

$$\phi_k(t+1) = \phi_k(t) + \alpha_v \left(r_k(t) + \gamma \vartheta_{\phi_k}(u_k(t+1)) - \vartheta_{\phi_k}(u_k(t))\right) \nabla_{\phi_k} \vartheta_{\phi_k}(u_k(t)) \tag{27}$$

where $\alpha_v$ is the learning rate for the critic network, and $\nabla_{\phi_k} \vartheta_{\phi_k}(u_k(t))$ is the gradient of the value function with respect to its parameters.

The global agent uses the DPPO approach to update its policies and value functions. The distribution center optimizes overall inventory management by coordinating actions across all retailers. The actor network for the distribution center determines the policy $\pi_{\theta_d}(u(t)|v(t))$ which dictates the quantities to order and distribute. The critic network evaluates the value function $\vartheta_{\phi_d}(u(t))$, which assesses the expected return from the global state $u(t)$. The policy update for the distribution center ensures stable updates by clipping the policy ratio as

$$\theta_d(t+1) = \theta_d(t) + \alpha_a \mathbb{E}\left[\min\left(\eta_t(\theta_d) \nabla_{\theta_d} \log \pi_{\theta_d}(v(t)|u(t)), clip(\eta_t(\theta_d), 1-\sigma, 1+\sigma).A_d^\pi\right)\right] \tag{28}$$

where $A_d^\pi = r(t) + \gamma \vartheta_{\phi_d}(u(t+1)) - \vartheta_{\phi_d}(u(t))$ represents advantage function for the distribution center, and $\eta_t(\theta_d)$ denotes is the probability ratio between the current and the old policy, calculated as

$$\eta_t(\theta_d) = \frac{\pi_{\theta_d}(u(t)|v(t))}{\pi_{\theta_d}^{old}(u(t)|v(t))} \tag{29}$$

If the ratio $\eta_t(\theta_d) > 1$, this means that new strategy is likely to be selected, otherwise the old strategy continues. This prevents new strategies from significantly deviating from old strategies by tailoring the optimization objective, while avoiding the over-estimation to the optimization objective with an adaptive divergence penalty coefficient. Also, the term $clip(\eta_t(\theta_d), 1-\sigma, 1+\sigma)$ is the clipped probability ratio to avoid large policy changes.

The value function update for the distribution center adjusts the parameters $\phi_d$ to minimize the temporal difference error as

$$\phi_d(t+1) = \phi_d(t) + \alpha_v \left(r(t) + \gamma \vartheta_{\phi_d}(u(t+1)) - \vartheta_{\phi_d}(u(t))\right) \nabla_{\phi_d} \vartheta_{\phi_d}(u(t)) \tag{30}$$

The global actor (distribution center) aggregates gradients from all local actors (retailers). With $k$ local retailers, the cumulative policy gradient for the global actor is calculated as

$$\Delta \theta_d = \frac{1}{K} \sum_{k=1}^{K} w_k \nabla_{\theta_k} \log \pi_{\theta_k}(v_k(t)|u_k(t)) A_k^\pi \tag{31}$$

where $\pi_{\theta_k}$ represents the policy of retailer $k$, parameterized by $\theta_k$ and $w_k = \mu_1 e^{-\rho(t-t_k)} + \mu_2 \frac{r_k}{\sum_{k=1}^{K} r_k}$ is an importance weighting factor where $e^{-\rho(t-t_k)}$ prioritizes recent updates from retailer $k$ and $\frac{r_k}{\sum_{k=1}^{K} r_k}$ ensures that the agent with high average rewards has high importance.

Similarly, the global critic aggregates the gradients from all local agents, and the cumulative value gradient for the global critic is

$$\Delta \phi_d = \frac{1}{K} \sum_{k=1}^{K} w_k \delta_t^k \nabla_{\phi_k} \vartheta_{\phi_k}(u_k(t)) \tag{32}$$

where $\vartheta_{\phi_k}(u_k(t))$ represents the value function parameterized by $\phi_k$ for retailer $k$ and $\delta_t^k$ is the temporal difference error, computed as

$$\delta_t^k = r_k(t) + \gamma \vartheta_{\phi_k}(u_k(t+1)) - \vartheta_{\phi_k}(u_k(t))) \tag{33}$$

where $r_k(t)$ is the reward by retailer $k$ at time $t$, $\vartheta_{\phi_k}(u_k(t+1))$ and $\vartheta_{\phi_k}(u_k(t))$ represents the estimated value of the next and current states, respectively.

To ensure stability and synchronization in the asynchronous learning process, exponential moving averages of the global parameters are incorporated as a stabilization mechanism as

$$\theta_d \leftarrow \tau \theta_d + (1-\tau)(\theta_d + \alpha_{a,d} \Delta \theta_d) \tag{34}$$

$$\phi_d \leftarrow \tau \phi_d + (1-\tau)(\phi_d + \alpha_{c,d} \Delta \phi_d) \tag{35}$$

where $\tau \in (0,1]$ controls update smoothness, $\alpha_{a,d}$ and $\alpha_{v,d}$ are the learning parameters of the global actor and critic network. After updating the global parameters, each local retailer receives the updated global parameters as

$$\theta_k \leftarrow \theta_d \text{ and } \phi_k \leftarrow \phi_d, \forall k \in \{1, 2, \ldots K\} \tag{36}$$

This ensures that the proposed algorithm effectively manages asynchronous gradient updates while maintaining stability and synchronization in the pharmaceutical supply chain scenario. The pseudocode summarizing the hybrid A3C-DPPO algorithm is outlined in Algorithm 1.

---

**Algorithm 1:** Cooperative A3C-DPPO algorithm for Agri-food Supply Chain

---

1: Initialize the parameters $\theta_d$ and $\phi_d$ for the global actor critic networks
2: Initialize the parameters $\theta_k$ and $\phi_k$ for the local actor critic networks for each retailer
3: Initialize the hyperparameters of actor and critic networks of local and global network
4: **for** each epoch:
5:    Reset the cumulative gradients for the global actor $(\Delta \theta_d)$ and global critic $(\Delta \phi_d)$
6:    **for** each local agent (retailer) $k$ in parallel:
7:       Simulate the local agent's interaction with the environment to collect state-action-reward-next state tuples
8:       Compute the advantage function $A_k^{\pi}$ for the local agent
9:       Calculate the local actor's policy gradient $\nabla_{\theta_k}$
10:      Compute the temporal difference error $\delta_t^k$ and the local critic's value gradient $\nabla_{\phi_k}$
11:      Accumulate these gradients into the global cumulative gradients $(\Delta \theta_d)$ and $(\Delta \phi_d)$
12:    **end for**
13:   Update the global actor and critic parameters using cumulative gradients.
14:   Synchronize the updated global parameters with all local agents.
15: **end for**
16: **end for**

# 5. Numerical Experiments

To evaluate the effectiveness of the proposed algorithm, experimental simulations are conducted for inventory management under demand and lead time uncertainties. In the agri-food supply chain scenario, the setup consists of a single agricultural farm and a single distribution center, which serves a varying number of retailers, forming a representative sub-network.. The experimental simulations are conducted on HP workstation with Core i9 processor, 32GB RAM, and RTX 3060 graphics card. The proposed A3C-DPPO algorithm is implemented with Python 3.6 and Pytorch tools and the performance is evaluated using real dataset. The hyper-parameters of A3C-DPPO algorithm are illustrated in Table 2.

Table 2. Hyperparameters of A3C-DPPO Algorithm

| Parameter | Value |
|---|---|
| Hidden layer size | 2 |
| Node size | {64, 128} |
| Learning rate (Actor Network) | 0.00005 |
| Learning rate (Critic Network) | 0.0001 |
| Clipping parameter $\sigma$ | [0.1, 0.2, 0.3] |
| Activation Function (Policy Network) | tanh |
| Activation Function (Value Network) | ReLU |

## A. Data and Parameters

This study considers three farm-fresh agricultural products—pomegranate (Product 1), bayberry (Product 2), and apple (Product 3)—which are supplied from the farm to distribution centers and subsequently distributed to multiple retailers. The inventory cost, production cost, unit sale price, and other inventory-related costs associated with each product throughout the supply chain are summarized in Tables 3 to 5.

Table 3 (a). Cost related parameters for Agricultural Farm

| Products | Unit Inventory cost ($) | Unit Production cost ($) |
|---|---|---|
| 1 | 0.04 | 3 |
| 2 | 0.12 | 8.2 |
| 3 | 0.14 | 9.5 |

Table 3 (b). Other transportation related parameters

| Parameter | Value | Parameter | Value |
|---|---|---|---|
| $M$ | 10 | $Q_m$ | 4000 units |
| $dist_{j,k}$ | 50 km | $c^m_{p,l}$ | $26 |
| $c^m_{p,un}$ | $26 | $c_{fuel}$ | 40/ltr |
| $c_{f,trans}$ | $18 | $c_f$ | $12 |

Table 4(a). Cost related parameters for Distribution Centers

| Products | Initial inventory | Unit purchase price ($) | Unit sale price ($) | Unit holding cost ($) | Unit shortage cost ($) | Unit wastage cost ($) | Fixed ordering price | Max. inventory capacity |
|---|---|---|---|---|---|---|---|---|
| p = 1 | 80 | 4.25 | 8.28 | 0.17 | 0.24 | 0.02 | 300 | 500 |
| p = 2 | 80 | 10.05 | 17 | 0.2 | 0.34 | 0.14 | 300 | 500 |
| p = 3 | 80 | 11.25 | 19.2 | 0.36 | 0.44 | 0.17 | 300 | 500 |

Table 5(a). Parameters for Retailer 1 associated with Distribution Center

| Products | Initial Inventory | Unit sales price ($) | Unit holding cost ($) | Unit shortage cost ($) | Unit wastage cost ($) | Fixed ordering price | Max. Inventory capacity | Demand Pattern |
|---|---|---|---|---|---|---|---|---|
| p = 1 | 60 | 25.8 | 0.2 | 0.32 | 0.04 | 200 | 200 | Normal |
| p = 2 | 20 | 35 | 0.28 | 0.38 | 0.16 | 200 | 200 | Weibull |
| p = 3 | 30 | 50 | 0.5 | 0.46 | 0.19 | 200 | 200 | Gamma |

Table 5(b). Parameters for Retailer 2 associated with Distribution Center

| Products | Initial Inventory | Unit sales price ($) | Unit holding cost ($) | Unit shortage cost ($) | Unit wastage cost ($) | Fixed ordering price | Max. Inventory capacity | Demand Pattern |
|---|---|---|---|---|---|---|---|---|
| p = 1 | 70 | 26 | 0.21 | 0.12 | 0.10 | 200 | 100 | Weibull |
| p = 2 | 30 | 35 | 0.27 | 0.18 | 0.18 | 200 | 100 | Normal |
| p = 3 | 50 | 52 | 0.43 | 0.38 | 0.2 | 200 | 100 | Gamma |

Table 5(c). Parameters for Retailer 3 associated with Distribution Center

| Products | Initial Inventory | Unit sales price ($) | Unit holding cost ($) | Unit shortage cost ($) | Unit wastage cost ($) | Fixed ordering price | Max. Inventory capacity | Demand Pattern |
|---|---|---|---|---|---|---|---|---|
| p = 1 | 60 | 27 | 0.19 | 0.14 | 0.09 | 200 | 100 | Gamma |
| p = 2 | 35 | 34 | 0.2 | 0.19 | 0.24 | 200 | 100 | Weibull |
| p = 3 | 42 | 50 | 0.41 | 0.39 | 0.28 | 200 | 100 | Normal |

For demand variation, the customer follows the mixture of Normal $\left(N(\mu,\sigma^2)\right)$ distributions. For instance, customer demand from Monday to Thursday follows a Normal distribution with the lowest mean, the demand on Friday follows a Normal distribution with the median mean, and demand during weekend follows a Normal distribution with the highest mean as illustrated in Table 6. Further, since customer demands are assumed to be independent and identically distributed (i.i.d) with inherent fluctuations, the Gamma distribution is utilized to capture the randomness and variability in demand patterns, characterized by a scale parameter $m$ and a shape parameter $u$. For experimental evaluation, variations in both the shape and scale parameters of the Gamma distribution $\left(Gamma(\alpha,\beta)\right)$ and Weibull distribution $\left(Weibull(u,\lambda)\right)$ are considered to reflect different levels of demand uncertainty as illustrated in Table 6.

Table 6 (a). Normal Distribution Demand patterns

| Days of the week | Mean | Demand Pattern |
|---|---|---|
| Mon-Wed | lowest | $N(3,1.5)$ |
| Thurs-Fri | medium | $N(6,1)$ |
| Sat-Sun | highest | $N(12,2)$ |

Table 6(b). Gamma and Weibull Distribution Demand patterns

| Days of the week | Demand Patterns | |
|---|---|---|
| Mon-Wed | $Gamma(2,10)$ | $(Weibull(1,0.5))$ |
| Thurs-Fri | $Gamma(4,5)$ | $(Weibull(3,3))$ |
| Sat-Sun | $Gamma(1,20)$ | $(Weibull(3,0.2))$ |

### B. Baseline Schemes

The performance of the proposed A3C-DPPO algorithm is compared with following baseline algorithms:

- $(s,S)$ *Policy*: A periodic-review $(s,S)$ policy as classical inventory policy is considered in which reorder when the inventory position drops below $s$ units, order need to be placed to replenish the inventory up to $S$ units for each period $t \in T$. Under this policy, the inventory position is reviewed in every $t$ time period. Here, $s$ is considered as the reorder point and $S$ referred as order-up-to-level.
    a) *Deterministic Demand Scenario:* In this case, the demand is assumed to follow a nominal distribution, and the corresponding parameter values are adopted as presented in [47].
    b) *Stochastic Demand Scenario*: For modeling demand uncertainty, an ellipsoidal uncertainty set is employed. The parameters defining the ellipsoidal distribution are tuned in accordance with the approach outlined in [47].

- *DQN Algorithm*: DQN policy is value-based policy to optimize the complex inventory operations. The hyperparameters of DQN is tuned in similar way as considered in [30].
- *SAC Algorithm*: To evaluate performance under stochastic environments with continuous action spaces, the SAC algorithm is considered. The associated hyperparameters are tuned according to the configuration presented in [30].
- *Centralized PPO:* The PPO algorithm is implemented in a centralized learning setting, where a global agent observes all environment states and actions, enabling policy updates through aggregated experiences [40].
- *Federated DRL*: A federated reinforcement learning approach is adopted to enable decentralized training across multiple agents. Local models are periodically synchronized with a global model to preserve privacy and reduce communication overhead, while still benefiting from shared policy improvement [48].

## C. Convergence Analysis

This section examines the convergence behavior of a proposed algorithm, likely designed for inventory management in an agri-food supply chain, by analyzing two key parameters: the discount factor and the clipping factor. The discount factor, illustrated in Fig. 3(a), determines the balance between immediate and future rewards in the decision-making process. The study found that higher discount factor values, particularly as γ approached 1, led to better performance. This suggests that the algorithm achieves optimal results when it places greater emphasis on long-term consequences, which is crucial in managing inventory for perishable goods and dealing with the seasonal variations typical in agri-food supply chains. Fig. 3(b) focuses on the clipping factor, a component of the A3C-DPPO algorithm, which limits how much the new policy can deviate from the old one during updates. The clipping factor values of 0.1, 0.2, and 0.3 are considered, and finding suggests that 0.2 yielded the best results. This middle value likely strikes an optimal balance between allowing for necessary policy adjustments and maintaining learning stability. The use of these parameters indicates that the algorithm aims to achieve a delicate equilibrium between making immediate inventory decisions and optimizing long-term supply chain performance, which is particularly important given the complex and volatile nature of agri-food supply chains.

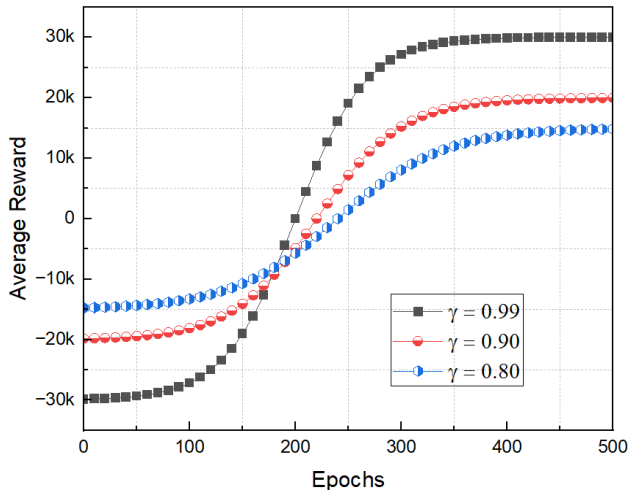

**Fig. 3(a)** Convergence behaviour under varying value of discount factor

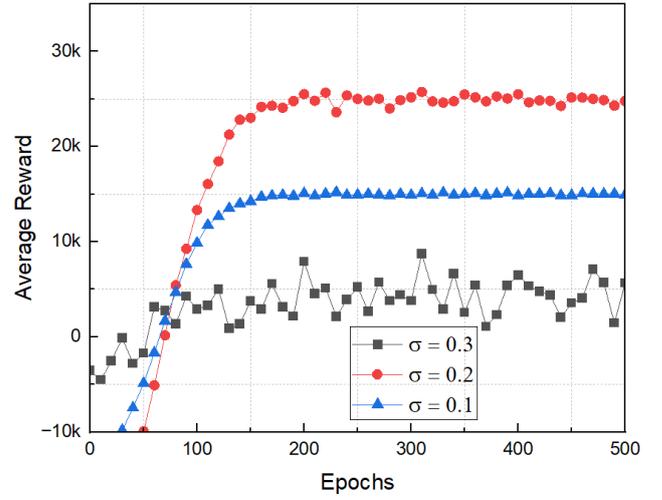

**Fig. 3(b)** Convergence behaviour under varying value of clipping factor

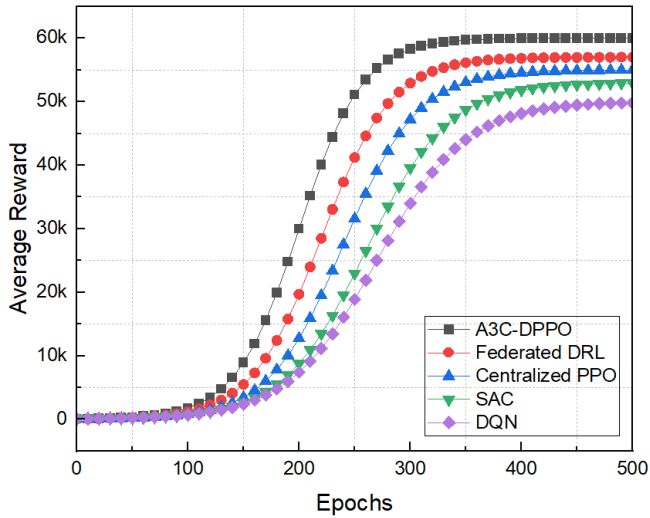

**Fig. 4(a)** Convergence comparison with baseline approaches

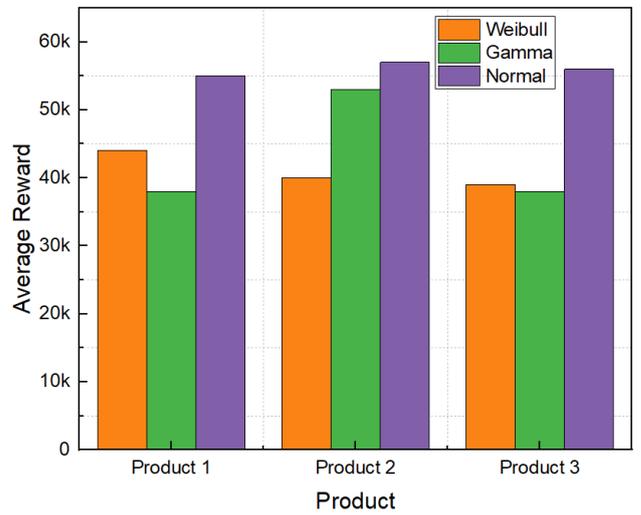

**Fig. 4(b)** Average reward under different demand scenarios

Additionally, to evaluate the learning efficiency and policy stability of the proposed method, the convergence behavior of A3C-DPPO against baseline approaches is compared, as shown in Fig. 4(a). The results demonstrate that A3C-DPPO achieves the highest average reward with the fastest convergence rate, stabilizing by around 300 epochs. This is attributed to its cooperative distributed architecture, which allows local agents to asynchronously update shared policies through parallel interactions with the environment. The shared gradient updates lead to faster adaptation and more sample-efficient learning, even under demand uncertainty and variable lead times. Federated DRL also performs well but converges more slowly due to its periodic model aggregation scheme and lack of inter-agent coordination. Centralized PPO converges moderately, hindered by its reliance on centralized sampling, which becomes less scalable in multi-agent environments. SAC, while effective in continuous control, shows slower learning progress and lower final reward values, due to its sensitivity to entropy regularization and exploration parameters. As illustrated, DQN exhibits the poorest performance, underscoring the limitations of value-based methods in high-dimensional continuous action spaces.

Table 7 and Fig. 4(b) illustrate the maximum overall profit achieved in the supply chain at the optimal replenishment level, considering various demand distributions. The results demonstrate that the proposed algorithm exhibits superior adaptability to the complex dynamics inherent in inventory management, particularly in the context of agri-food supply chains. Specifically, the algorithm shows enhanced responsiveness to demand uncertainties, seasonal fluctuations, and other dynamic factors such as product perishability and variable lead times. The strength of the proposed approach lies in its ability to efficiently manage large-scale, complex supply chain environments. This is achieved through the implementation of parallel agents, each optimizing different aspects of the inventory system simultaneously. This parallel processing capability allows for a more comprehensive and nuanced optimization of the entire supply chain, rather than focusing on single entity in a supply chain. The performance across different demand distributions underscores robustness and versatility of proposed algorithm. By effectively handling various demand scenarios, the system can maintain optimal inventory levels and maximize profitability, even in the face of unpredictable market conditions. This adaptability is particularly crucial in the agri-food sector, where demand can be highly variable due to factors such as changing consumer preferences, weather conditions, and seasonal trends.

Table 7. Total profit obtained with proposed algorithm in supply chain

| Demand Distribution | Manufacturer | Distribution Center | Retailer | Supply chain system |
|---|---|---|---|---|
| **Product 1** | | | | |
| Weibull | 59,656 | 1,823 | 1,173 | 62,652 |
| Gamma | 56,322 | 1,652 | 998 | 58,972 |
| **Product 2** | | | | |
| Weibull | 58,980 | 1,225 | 1,000 | 61,205 |
| Gamma | 60,151 | 1,532 | 1,144 | 62,827 |
| **Product 3** | | | | |
| Weibull | 57,443 | 1,342 | 1,298 | 60,083 |
| Gamma | 59,662 | 1,409 | 1,287 | 62,358 |

Note. The profit units are measured in USD

### D. Performance Comparison with Baseline Schemes

In this sub-section, the performance of proposed A3C-DPPO based inventory policy with baseline algorithms has been compared. As illustrated from Fig. 5, the proposed Cooperative A3C-DPPO framework consistently outperforms all baselines, demonstrating both superior learning capacity and resilience under stochastic conditions. At low demand variance (0%), A3C-DPPO representing an improvement of 21.7% over DQN, 5.7% over SAC, 7.7% over centralized PPO, and 3.7% over Federated DRL. As demand variance increases to 80%, A3C-DPPO outperforming DQN by 220.0%, SAC by 33.3%, centralized PPO by 18.5%, and Federated DRL by 6.7%. These findings highlight the robustness of the A3C-DPPO framework in uncertain environments, where state-of-the-art DRL methods suffer sharp degradation. The distributed training approach of proposed A3C-DPPO results in its improved performance over DRL algorithms, resulting in a more stable and consistent learning curve. Fig. 6 depicts the performance of proposed algorithm under stochastic lead time scenarios. The lead time uncertainty is mapped using Exponential and Gamma Distribution representing different demand distribution scenarios. The scatter plot shows how inventory cost varies with lead time across four scenarios. The trend lines for each scenario (dashed lines) indicate that as lead time increases, the inventory cost generally rises. It suggests that longer lead times are associated with higher inventory costs. The wide scatter of points also suggests significant variability in costs for any given lead time, highlighting the complexity and unpredictability in supply chain management. Additionally, the proposed algorithm demonstrates robustness by effectively handling slight variations in replenishment lead time, thereby adapting well to different lead time situations. In addition, Table 8 illustrates the performance comparison under varying lead time scenarios with varying value of $\lambda$ as a parameter of Exponential distribution, to model the changes in lead time. As $\lambda$ increases, the average lead time decreases. The proposed algorithm demonstrates lowest cost across all scenarios while DQN typically shows highest costs, especially for longer lead times. As observed, A3C-DPPO outperforms the other strategies consistently because it adept at adapting to these changing conditions through more sophisticated learning algorithms to handle the dynamic nature of inventory management.

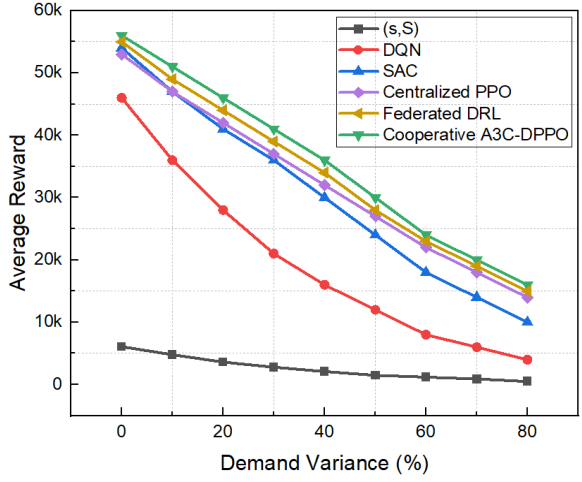 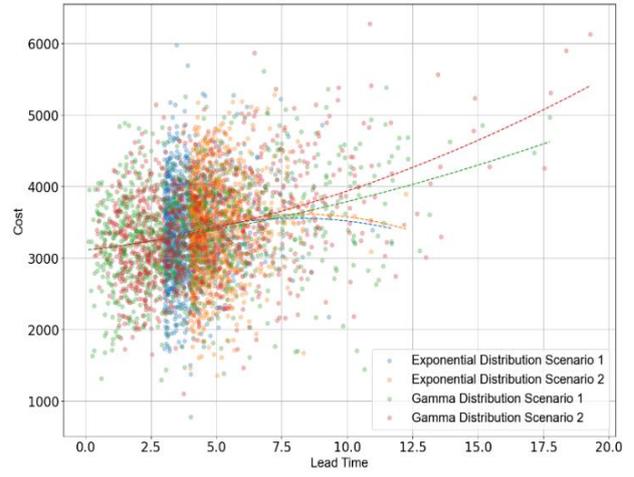

**Fig. 5** Performance comparison under different demand variance  **Fig. 6** Performance of A3C-DPPO algorithm under different lead time distribution

Table 8. Cost comparison under varying lead time

| $\lambda$ | Average Lead Time | DQN | Optimality Gap (%) | SAC | Optimality Gap (%) | Centralized PPO | Optimality Gap (%) | Federated DRL | Optimality Gap (%) | Cooperative A3C-DPPO |
|---|---|---|---|---|---|---|---|---|---|---|
| 0.1 | 10 days | 6872 ± 95 | 40.5% | 5564 ± 82 | 13.6% | 4998 ± 70 | 2.0% | 4960 ± 65 | 1.3% | 4898 ± 52 |
| 0.5 | 2 days | 5440 ± 88 | 25.8% | 4986 ± 74 | 15.3% | 4455 ± 63 | 3.0% | 4388 ± 58 | 1.5% | 4324 ± 49 |
| 1.0 | 1 day | 4876 ± 76 | 25.0% | 4265 ± 62 | 9.4% | 3980 ± 55 | 2.1% | 3945 ± 51 | 1.2% | 3900 ± 47 |
| 1.5 | 0.67 day | 4198 ± 70 | 23.8% | 3782 ± 58 | 11.5% | 3560 ± 50 | 5.1% | 3458 ± 44 | 2.0% | 3392 ± 42 |
| 2.0 | 0.5 day | 3870 ± 62 | 61.6% | 2900 ± 47 | 21.0% | 2600 ± 40 | 8.4% | 2498 ± 38 | 4.2% | 2398 ± 35 |

Furthermore, Fig.7 depicts the variation in average inventory cost across baseline schemes under different lead time scenarios. As seen from figure, the proposed cooperative A3C-DPPO algorithm consistently incurs the lowest cost across all lead time brackets, outperforming baseline methods in both short and long-time scenarios. Specifically, in the 0–2-day range, A3C-DPPO achieves the lowest inventory cost among all methods, with a improvement of approximately 6-10% over SAC and Federated DRL. As lead time increases to 6-8 days, A3C-DPPO incurs ~8% lower cost compared to centralized PPO and ~10% lower than DQN. These results underscore the effectiveness of cooperative A3C-DPPO in capturing both local and global dynamics within the supply chain, leveraging parallel learning agents and cooperative gradient sharing to mitigate lead time variability. Unlike DQN, which struggles with delayed feedback, the proposed method adapts well to increasing uncertainty.

Fig. 8 illustrates the convergence time measured in training iterations. It is observed that as the lead time increases, the convergence time increases. Among all algorithms, the proposed algorithm consistently achieves the fastest convergence across all lead time scenarios. As lead times extend to the most challenging 8+ day scenario, A3C-DPPO still converges in under 170 iterations, outperforming SAC, Centralized PPO, Federated DRL, and DQN. The superior convergence behavior of A3C-DPPO can be attributed to its cooperative gradient-sharing mechanism and parallelized actor-critic updates, which enhance sample efficiency and learning stability. In contrast, value-based methods like DQN suffer more significantly from long lead time delays, as delayed rewards hinder stable Q-value estimation.

To validate the scalability and effectiveness of the proposed cooperative framework, the supply chain networks comprising 3 farms, 4 DCs, and 10 retailers is setup. Each farm produces different product categories with varied perishability and lead time profiles. As seen in Table 9, proposed algorithm achieved the highest average cumulative reward outperforming federated DRL, centralized PPL, and SAC.

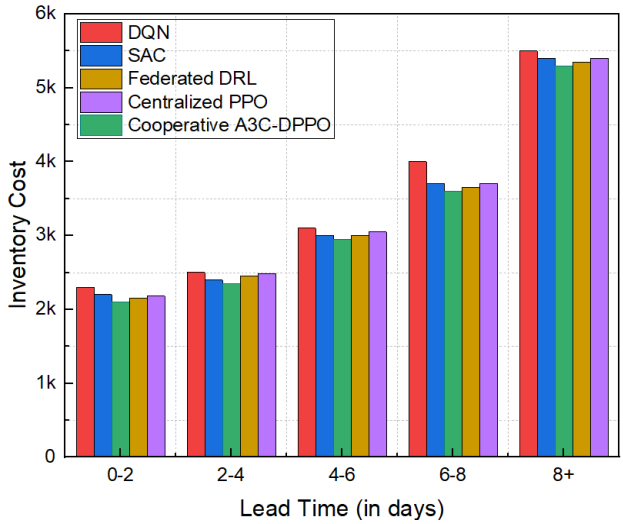
Fig. 7 Performance comparison under different lead time scenarios

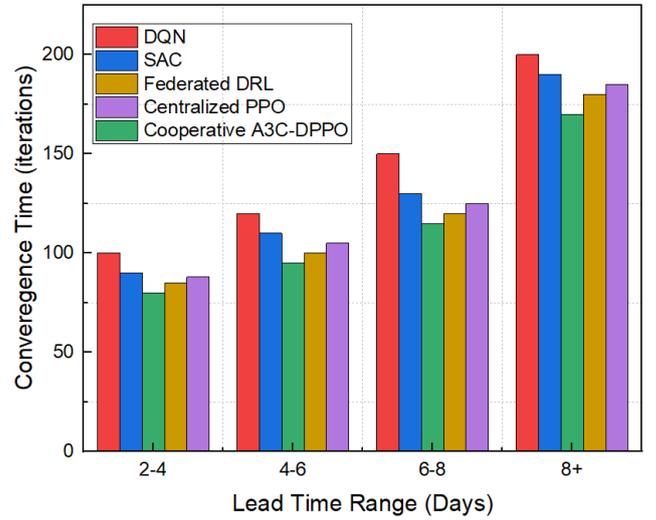
Fig 8. Convergence time across different lead time scenarios

Additionally, A3C-DPPO required the fewest training iterations to converge compared to other baselines. The results demonstrates that the cooperative distributed learning mechanism of A3C-DPPO maintains performance advantages even in larger and more complex networks. In contrast, centralized and federated baselines exhibit higher convergence delays and cost inefficiencies. The experiment confirms that A3C-DPPO generalizes effectively to multi-echelon, agent-dense environments, validating its scalability and robustness claims.

Table 9. Performance comparison under large supply chain network

| Method | Average Reward | Average Inventory Cost | Convergence (iterations) |
|---|---|---|---|
| SAC | 52,800 | 17,300 | 190 |
| Centralized PPO | 56,100 | 16,000 | 165 |
| Federated DRL | 58,300 | 15,200 | 150 |
| Cooperative A3C-DPPO | 61,200 | 14,500 | 135 |

*E. Sensitivity Analysis*

Fig. 9 presents the sensitivity of average inventory cost components under varying perishability rates $(\delta)$. As $\delta$ increases from 0.05 to 0.20, all cost components show upward trends, with the wastage penalty experiencing the most significant surge. This validates the impact of perishability modeling in agri-food supply chains. The surge in wastage cost emphasizes the importance of integrating freshness-aware decision-making, as enabled by proposed A3C-DPPO framework. Additionally, transportation and stockout costs also rise due to the need for more frequent deliveries and conservative inventory holding policies, respectively. Overall, the analysis confirms the robustness and adaptability of proposed cooperative A3C-DPPO model under high perishability stress.

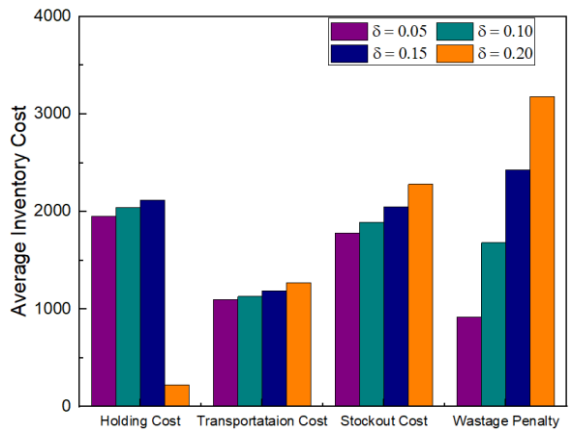
Fig. 9 Impact of Perishability Rate $(\delta)$ on Inventory Cost Components

6. **Theoretical Contributions and Managerial Implications**

*6.1 Theoretical Contributions*

This study makes significant contributions to the theoretical understanding of the agri-food supply chain management through the introduction of a novel A3C-DPPO algorithm. This novel approach is specifically designed to address the unique challenges faced in the agri-food supply chain networks, with a particular focus on managing uncertainties stemming from stochastic demand and variable replenishment lead times. The proposed algorithm advances existing research in several key ways. Firstly, the framework effectively handles the inherent uncertainties in both demand and lead times, which are critical factors influencing agri-food supply chain performance. By modeling these uncertainties explicitly, the framework enables more accurate and stable decision-making under volatile market conditions. Second, the algorithm emphasizes the optimization of dynamic inventory policies, recognizing their strategic importance in uncertainty mitigation and profit maximization. This approach allows for more nuanced and effective inventory management strategies, extending prior works that has largely focused on either centralized or non-cooperative paradigms.

*6.2 Managerial Implications*

From a practical standpoint, the study offers several valuable insights and tools for supply chain managers operating in uncertain and dynamic environments.

- *Enhanced Scalability and Efficiency:* The proposed cooperative A3C-DPPO framework improves computational performance by enabling parallelized training across agents. This allows model to scale effectively with network size, accommodating larger or more complex agri-food supply chains without compromising on learning quality.
- *Improved Responsiveness and Resilience:* By optimizing replenishment decisions, the algorithm enhances the supply chain's ability to respond to uncertainties and maintain resilience in the face of unexpected changes. Such responsiveness contributes to lower inventory costs, improved fill rates, and stronger resilience against supply disruptions or seasonal variability.
- *Continuous Learning and Adaptation:* The algorithm facilitates real-time learning by allowing distribution centers and retailers to act as intelligent agents. These agents adapt their ordering strategies as environmental dynamics shift, enabling a self-improving system capable of maintaining performance even in highly variable conditions.
- *Decision Support for Inventory Managers:* The findings provide inventory managers with actionable insights into optimal inventory policies under uncertainty. It helps managers to strike a balance between service levels targets, cost efficiency, and freshness losses with responsive policy formulation. This ultimately contributes to better customer service, improved operational resilience and long-term competitiveness.

*6.3 Alignment with SDGs*

This research aligns closely with Sustainable Development Goal 12 - "Responsible Production and Consumption." By focusing on effective inventory management under uncertainties in the agri-food supply chain, the study contributes to more sustainable practices in several ways:

- *Reduction of Food Waste:* Optimized inventory management can lead to reduced spoilage and waste of perishable food products, contributing to more sustainable consumption patterns
- *Resource Efficiency:* By improving the accuracy of demand forecasting and inventory decisions, the algorithm helps minimize overproduction and excess inventory, leading to more efficient use of resources throughout the supply chain.
- *Economic Sustainability:* The enhanced profitability and resilience resulting from improved inventory management contribute to the economic sustainability of businesses in the agri-food sector.
- *Adaptive Capacity:* The continuous learning aspect of the algorithm supports the development of more adaptable and resilient supply chains, which is crucial for long-term sustainability in the face of climate change and other global challenges.

In summary, this study not only advances the theoretical understanding of agri-food supply chain management but also provides practical tools and insights that can drive significant improvements in real-world operations. By addressing key challenges such as uncertainty, perishability, and the need for adaptive strategies, the research contributes to both the economic efficiency and environmental sustainability of agri-food supply chains, aligning with broader global sustainability goals.

## 7. Conclusion

The uncertainty associated with agri-food supply chain has significant impact on supply chain decisions, highlighting the need of robust framework to optimize inventory decisions. This study addresses this challenge by proposing a novel algorithm designed to optimize inventory management under market uncertainties, including stochastic customer demand and variable replenishment lead times. The A3C-DPPO algorithm to recommend near-optimal inventory replenishment policy by continuously adapting the dynamic and complex environment of the underlying supply chain. By selecting optimal ordering quantities in a continuous action space, the proposed method effectively handles uncertainty in both demand and supply. The effectiveness of proposed algorithm has been investigated under stochastic demand scenarios. The computational results show the superior performance of proposed inventory policy to DQN based inventory policy as well as classical *(s, S)* policy in terms of reducing total inventory cost and improving overall profit of agri-food supply chain. It is also demonstrated with numerical results that A3C-DPPO based inventory policy is relatively stable even under highly dynamic and stochastic market uncertainties, where classical *(s, S)* policy reflects highly volatile nature. The proposed model provides decision-makers in the food industry with a valuable tool to optimize inventory operations to improve its overall resilience, responsiveness and efficiency of food supply chain. Although the proposed research encompasses various aspects to handle uncertainties in agri-food supply chain, but has few limitations. Firstly, the simplifying assumptions to manage the complexity of the system, but this leads to over-simplified real-world dynamics and thus leads to potential biases and inaccuracies. Secondly, the social and environmental sustainability aspects is not considered in this research, that are essential to improve the overall performance of supply chain by optimizing transportation routes, reducing $CO_2$ emission and emphasizing long-term sustainability by balancing profit objectives with environmental impact.

Building upon the current research, there are compelling avenues for future exploration that could significantly enhance the efficiency and effectiveness of agri-food supply chains. A primary area of interest is the integration of food cold chain requirements, which are vital for maintaining the quality and safety of fresh agricultural products during their transportation through the supply chain. This extension could involve optimizing temperature-controlled logistics, assessing the impact of cold chain breaches, and analyzing the cost-effectiveness of advanced cold chain technologies. Simultaneously, investigating the role of cross-docking facilities as a critical logistical component presents another promising direction. These facilities have the potential to optimize efficiency and reduce inventory holding costs by minimizing processing time and shipment expenses. Future studies could focus on the optimal placement and design of cross-docking facilities, developing efficient scheduling algorithms, and integrating cross-docking strategies with the proposed A3C-DPPO algorithm. Moreover, exploring multi-sourcing and cross-connection strategies among distribution centers and retailers could improve supply chain robustness and flexibility. By combining these aspects, the researchers could uncover valuable insights into creating a more responsive, efficient, and sustainable agri-food supply chain, ultimately leads to reduced costs, and enhanced overall profitability.